\documentclass{article}

    \PassOptionsToPackage{numbers, sort, compress}{natbib}
\usepackage[preprint]{neurips_2024}

\usepackage[utf8]{inputenc} % allow utf-8 input
\usepackage[T1]{fontenc}    % use 8-bit T1 fonts
\usepackage{hyperref}       % hyperlinks
\usepackage{url}            % simple URL typesetting
\usepackage{booktabs}       % professional-quality tables
\usepackage{amsfonts}       % blackboard math symbols
\usepackage{nicefrac}       % compact symbols for 1/2, etc.
\usepackage{microtype}      % microtypography
\usepackage{xcolor}         % colors

\usepackage{graphicx}
\usepackage{algorithm, algorithmicx, algpseudocode}
\usepackage{array}
\usepackage{multicol}
\usepackage{multirow}
\usepackage{svg}
\usepackage{soul}
\usepackage{caption}
\usepackage{subcaption}
\usepackage{amsmath,amsfonts}
\usepackage{makecell} % for multi-line cells

\algrenewcommand\algorithmicrequire{\textbf{Input:}}
\algrenewcommand\algorithmicensure{\textbf{Output:}}

\title{Venomancer: Towards Imperceptible and Target-on-Demand Backdoor Attacks in Federated Learning}

\author{%
  Son Nguyen$^{1}$, Thinh Nguyen$^{2}$, Khoa D Doan$^{1, 2}$, Kok-Seng Wong$^{1, 2}$\\
  $^1$College of Engineering \& Computer Science, VinUniversity, Hanoi, Vietnam\\
  $^2$VinUni-Illinois Smart Health Center, VinUniversity, Hanoi, Vietnam\\
  \texttt{\{son.nh, thinh.nth, khoa.dd, wong.ks\}@vinuni.edu.vn}
}

\begin{document}

\maketitle

\begin{abstract}
  Federated Learning (FL) is a distributed machine learning approach that maintains data privacy by training on decentralized data sources. Similar to centralized machine learning, FL is also susceptible to backdoor attacks, where an attacker can compromise some clients by injecting a backdoor trigger into local models of those clients, leading to the global model's behavior being manipulated as desired by the attacker. Most backdoor attacks in FL assume a predefined target class and require control over a large number of clients or knowledge of benign clients' information. Furthermore, they are not imperceptible and are easily detected by human inspection due to clear artifacts left on the poison data. To overcome these challenges, we propose \textbf{Venomancer}, an effective backdoor attack that is imperceptible and allows target-on-demand. Specifically, imperceptibility is achieved by using a visual loss function to make the poison data visually indistinguishable from the original data. Target-on-demand property allows the attacker to choose arbitrary target classes via conditional adversarial training. Additionally, experiments showed that the method is robust against state-of-the-art defenses such as Norm Clipping, Weak DP, Krum, Multi-Krum, RLR, FedRAD, Deepsight, and RFLBAT. The source code is available at \url{https://github.com/nguyenhongson1902/Venomancer}.
\end{abstract}

\section{Introduction}
\label{introduction}

In recent years, Federated Learning (FL)~\citep{mcmahan2017communication} emerged as a promising distributed machine learning paradigm, where multiple clients, i.e. smartphones or edge devices, collaboratively train a global model without sharing their private data. FL has been widely adopted in various applications such as mobile keyboard prediction~\citep{hard2018federated}, keyword spotting~\citep{leroy2019federated}, and healthcare informatics~\citep{xu2021federated}. Despite its collaborative training capability, many works~\citep{bagdasaryan2020backdoor, fang2023vulnerability, zhang2024a3fl, xie2019dba, zhang2022neurotoxin, wang2020attack} have shown that FL is vulnerable to various backdoor attacks, where an attacker can compromise some clients to inject a backdoor trigger into local models of those clients, leading to the global model's behavior being manipulated as desired by the attacker. Specifically, the backdoored global model will behave normally on the clean test data but misclassify the poisoned test data as the attacker-chosen class, e.g. a car's camera may recognize a stop sign patched with a sticky note on the road as a limit sign.
% i.e. the test data patched with a trigger, as the attacker-chosen class.

Currently, there are two main types of backdoor attacks in FL~\citep{zhang2024a3fl}, i.e. \textit{fixed-pattern attacks}~\citep{bagdasaryan2020backdoor, xie2019dba, zhang2022neurotoxin, wang2020attack} and \textit{trigger-optimization attacks}~\citep{fang2023vulnerability, zhang2024a3fl}, depending on how the trigger is created. Note that in some works, trigger-optimization attacks are also referred to as sample-specific attacks. For fixed-pattern attacks, the attacker uses a predefined trigger pattern to create the poisoned data.  While this type of attack is straightforward, it often suffers from performance degradation. In contrast, trigger-optimization attacks optimize the trigger to improve the success of the attacks. Fang et al.~\citep{fang2023vulnerability} proposed a new attack, F3BA, which enhances backdoor injection by selectively flipping the signs of a small proportion of model weights and jointly optimizing the trigger pattern with client models. A3FL~\citep{zhang2024a3fl}, which adversarially adapts the backdoor trigger to make it less likely to be removed by the global training dynamics, i.e. the server can train the global model to unlearn the trigger.

Although both types of backdoor attacks in FL can be effective against certain defenses, they still have some limitations. First, most existing backdoor attacks in FL leave clear artifacts on the poisoned data that can be easily detected when human inspection is involved (see Figure~\ref{fig:triggers_venom_a3fl}). In general, some studies~\citep{luo2018towards, Doan_2021_ICCV} aim to achieve the imperceptibility of the backdoor attack in centralized settings by clamping the perturbation within a very small range. Chen et al.~\citep{chen2017targeted} proposed Blend attack, which makes the trigger invisible by using a blend ratio. In FL, IBA~\citep{iba} also applies a small upper bound on the trigger to make it visually imperceptible. However, we show that the conventional way of achieving imperceptibility is not effective in FL since the FL backdoor attack is more challenging due to the decentralized nature (non i.i.d) of the data (see Figure~\ref{fig:eps_approach}). Additionally, backdoor attacks in FL are more difficult because the local updates of benign clients outnumber malicious updates, thus such benign updates can dilute the effect of the malicious ones after multiple rounds of aggregation at the server and make the backdoor attack less effective. Second, the attack requires prior knowledge of the target class before the training and the attacker cannot modify the target during inference.
% Additionally, the main thing that makes the backdoor attack in FL more challenging is that local updates of benign clients outnumber malicious updates, thus such benign updates can dilute the effect of the malicious ones after multiple rounds of aggregation at the server and make the backdoor attack less effective. Second, the attack requires prior knowledge of the target class before the training and the attacker cannot modify the target during inference.
% Second, the attacker needs to know the target class before training and is not able to change the target during inference, which limits the attack's flexibility.

Motivated by these challenges, in this paper, we propose \textbf{Venomancer}, a novel and effective backdoor attack in FL that adaptively optimizes the trigger pattern to make the poisoned data imperceptible to human beings and allows the attacker to select arbitrary target class at inference (i.e. target-on-demand property). Specifically, we achieve imperceptibility by utilizing a visual loss to make the poisoned data visually indistinguishable from the original data. We also allow the target-on-demand property during inference time by using conditional adversarial training. Our attack is a two-stage scheme, (1) training generator and (2) injecting backdoor. In the first stage, the generator is trained to generate adversarial noise acting as a trigger to create the poisoned data. In the second stage, the poisoned data is injected into the local model to train the backdoor. We further show that our attack only requires a small number of malicious or compromised clients (i.e. 2\% of the total clients) to achieve high accuracy on both the main task and the backdoor task. To the best of our knowledge, this is the first work that investigates imperceptible and target-on-demand backdoor attacks in FL.

% Move to related work
% Because the naive attack in~\citep{bagdasaryan2020backdoor} using a fixed trigger does not work effectively in FL, Bagdasaryan et al.~\citep{bagdasaryan2020backdoor} first proposed a model poisoning attack, where the backdoor can be effectively implemented by replacing the global model with the attacker's malicious models through carefully scaling model updates sent to the server for aggregation.

Our main contributions to the paper are summarized as follows:
\begin{itemize}
    \item We propose \textbf{Venomancer}, a novel and effective FL backdoor attack that is imperceptible to human beings and allows the target-on-demand property, meaning that the attacker can choose arbitrary target classes during inference. This attack significantly increases the stealthiness and flexibility within the FL setting.
    \item We propose a two-stage attack framework, i.e. (1) training generator and (2) injecting backdoor, that alternately optimizes the generator and the local model to achieve the imperceptibility and target-on-demand properties. Hence, the attack can effectively inject the backdoor into the global model with a small number of compromised clients.
    \item We empirically evaluate the effectiveness of our attack on multiple benchmark datasets, including MNIST~\citep{deng2012mnist}, Fashion-MNIST~\citep{xiao2017fashion} (F-MNIST), CIFAR-10~\citep{krizhevsky2009learning}, and CIFAR-100~\citep{krizhevsky2009learning}.
    \item We extensively demonstrate the robustness of our attack against state-of-the-art defenses, including Norm Clipping~\citep{sun2019can}, Weak DP~\citep{sun2019can}, Krum~\citep{blanchard2017machine}, Multi-Krum~\citep{blanchard2017machine}, RLR~\citep{ozdayi2021defending}, FedRAD~\cite{sturluson2021fedrad}, Deepsight~\citep{rieger2022deepsight}, and RFLBAT~\citep{wang2022rflbat}.
\end{itemize}

The rest of the paper is structured as follows. In Section~\ref{sec:related}, we discuss the existing backdoor attacks and defenses in FL. The threat model is outlined in Section~\ref{sec:threat}. We provide the details of the proposed method in Section~\ref{sec:method} and evaluate its performance in Section~\ref{sec:exps}. Section~\ref{sec:limitation} discusses the limitations of the attack, and Section~\ref{sec:conclusion} provides our conclusions. Further information regarding experimental settings and results can be found in the appendix.

\section{Preliminaries and related work}
\label{sec:related}

\subsection{Federated Learning}
Federated Learning (FL) was initially introduced in~\citep{mcmahan2017communication} to enhance data privacy and communication efficiency within decentralized learning environments. Formally, the central server initiates each communication round by randomly selecting a subset of $M$ clients from a total pool of $N$ clients to contribute to the training process. Every chosen client $i$ possesses a unique local dataset, denoted as $\mathcal{D}_{i}$. At the start of the round, the server dispatches the global model's parameters $\theta^{t-1}_{global}$ from the previous round to these selected clients. Each client, upon receiving these parameters, initializes its local model $f_i$ with $\theta^{t-1}_{global}$ and proceeds to train $f_i$ on its $\mathcal{D}_{i}$ across several local epochs. This process yields updated local model weights $\theta^{t}_{i}$, as outlined in Appendix~\ref{app:benign}. Subsequently, each client sends the difference $\left ( \theta^{t}_{i} - \theta^{t-1}_{global} \right )$ back to the central server. The server then aggregates these differences from all participating clients to update the global model's parameters to $\theta^{t}_{global}$ using the FedAvg~\citep{mcmahan2017communication} algorithm with the following equation:
% $\theta^{t}_{global} = \theta^{t-1}_{global} + \sum_{i=1}^{M}\frac{n_i}{n^t}\left ( \theta^{t}_{i} - \theta^{t-1}_{global} \right )$, where $n_i$ is the number of samples from client $i$, and $n^t$ is the total number of samples from the selected clients at round $t$. The FL training is repeated for a specific number of communication rounds. The detailed algorithm for local training of a benign client can be found in Appendix~\ref{app:benign}.
\begin{equation}
    \theta^{t}_{global} = \theta^{t-1}_{global} + \sum_{i=1}^{M}\frac{n_i}{n^t}\left ( \theta^{t}_{i} - \theta^{t-1}_{global} \right ) 
\end{equation}
where $n_i$ is the number of samples from client $i$, and $n^t$ is the total number of samples from the selected clients at round $t$. The FL training is repeated for a specific number of communication rounds. The detailed algorithm for local training of a benign client can be found in Appendix~\ref{app:benign}.

% \begin{equation} 
%   \label{eq:1}
%   \theta^{t}_{global} = \theta^{t-1}_{global} + \sum_{i=1}^{m}\frac{n_i}{n^t}\left ( \theta^{t}_{i} - \theta^{t-1}_{global} \right ),
% \end{equation}
% where $n_i$ is the number of samples from client $i$, and $n^t$ is the total number of samples from the selected clients at round $t$. The FL training is repeated for a specific number of communication rounds.

\subsection{Existing backdoor attacks in FL}
Because of the decentralized nature, FL opens a new surface for various backdoor attacks, where an attacker aims to inject a backdoor into the global model via local updates of malicious clients. Existing backdoor attacks in FL can be categorized into two types: \textit{fixed-pattern attacks}~\citep{bagdasaryan2020backdoor, xie2019dba, zhang2022neurotoxin, wang2020attack} and \textit{trigger-optimization attacks}~\citep{fang2023vulnerability, zhang2024a3fl, iba}. Because the naive attack in~\citep{bagdasaryan2020backdoor} using a fixed trigger does not work in FL, Bagdasaryan et al.~\citep{bagdasaryan2020backdoor} first proposed a model replacement attack, where the backdoor can be effectively implemented by replacing the global model with the attacker's malicious models through carefully scaling model updates sent to the server for aggregation. Xie et al.~\citep{xie2019dba} introduced the Distributed Backdoor Attack (DBA), which splits the global trigger into multiple local triggers for poisoning. Wang et al.~\citep{wang2020attack} proposed an edge-case backdoor attack, which strategically uses the out-of-distribution dataset inserted at the tails of the original data distribution, making them less prominent in the clients' training dataset. Neurotoxin~\citep{zhang2022neurotoxin} selects dimensions of parameters that are less likely to be updated by benign clients to inject the backdoor using those dimensions, thus making the backdoor more durable. Similarly, F3BA~\citep{fang2023vulnerability} projects gradients to infrequently updated model parameters like Neurotoxin. F3BA also optimizes the trigger pattern to maximize the difference between latent representations of clean and poisoned data. Nguyen et al.~\citep{iba} proposed IBA, which leverages the updated history of malicious clients' parameters to make the backdoor more persistent. A3FL~\citep{zhang2024a3fl} adapts the backdoor trigger to make it less likely to be removed by the global training dynamics.

\subsection{Existing defenses in FL}
There have been many studies on possible defenses that could mitigate backdoor attacks in FL. Based on the mechanisms, existing defenses in FL can be categorized into one of these types: \textit{norm clipping}~\citep{sun2019can}, \textit{deep model inspection}~\citep{rieger2022deepsight}, \textit{dimensionality reduction}~\citep{wang2022rflbat}, \textit{model-refinement}~\citep{sturluson2021fedrad}, and \textit{robust aggregation}~\citep{blanchard2017machine, ozdayi2021defending}. Norm Clipping~\citep{sun2019can} (NC) prevents large updates from malicious clients from disproportionately influencing the global model. Inspired by~\citep{ma2019data, dwork2006calibrating, abadi2016deep}, Weak Differential Privacy~\citep{sun2019can} (Weak DP) is not designed for privacy but to prevent backdoor attacks by perturbing a small amount of Gaussian noise to the malicious updates after norm clipping, making it harder for the compromised clients to inject backdoor without detection. Deepsight~\citep{rieger2022deepsight} performs deep model inspection for each model by checking its Normalized Update Energies (NEUPs), Division Differences (DDifs), and Threshold Exceedings (TEs). RFLBAT~\citep{wang2022rflbat} leverages Principal Component Analysis (PCA) and K-means~\citep{wu2012advances} to filter out malicious updates by identifying and clustering similar gradients in the low-dimensional space, allowing the aggregation of only benign updates. FedRAD~\citep{sturluson2021fedrad} uses a median-based scoring system along with knowledge distillation to filter out malicious updates and aggregate local updates effectively. Krum/Multi-Krum~\citep{blanchard2017machine} filters out the client with the smallest pairwise distance from other clients and trains the global model only with the filtered client updates. Ozdayi et al.~\citep{ozdayi2021defending} proposed the Robust Learning Rate (RLR), which dynamically adjusts the learning rate of the aggregation server based on the sign of the updates received from the clients, ensuring that updates leading to backdoor activations are minimized.

\section{Threat model}
\label{sec:threat}
\subsection{Attacker's goals}
\begin{figure}[t]
  \centering
  \includegraphics[width=\linewidth]{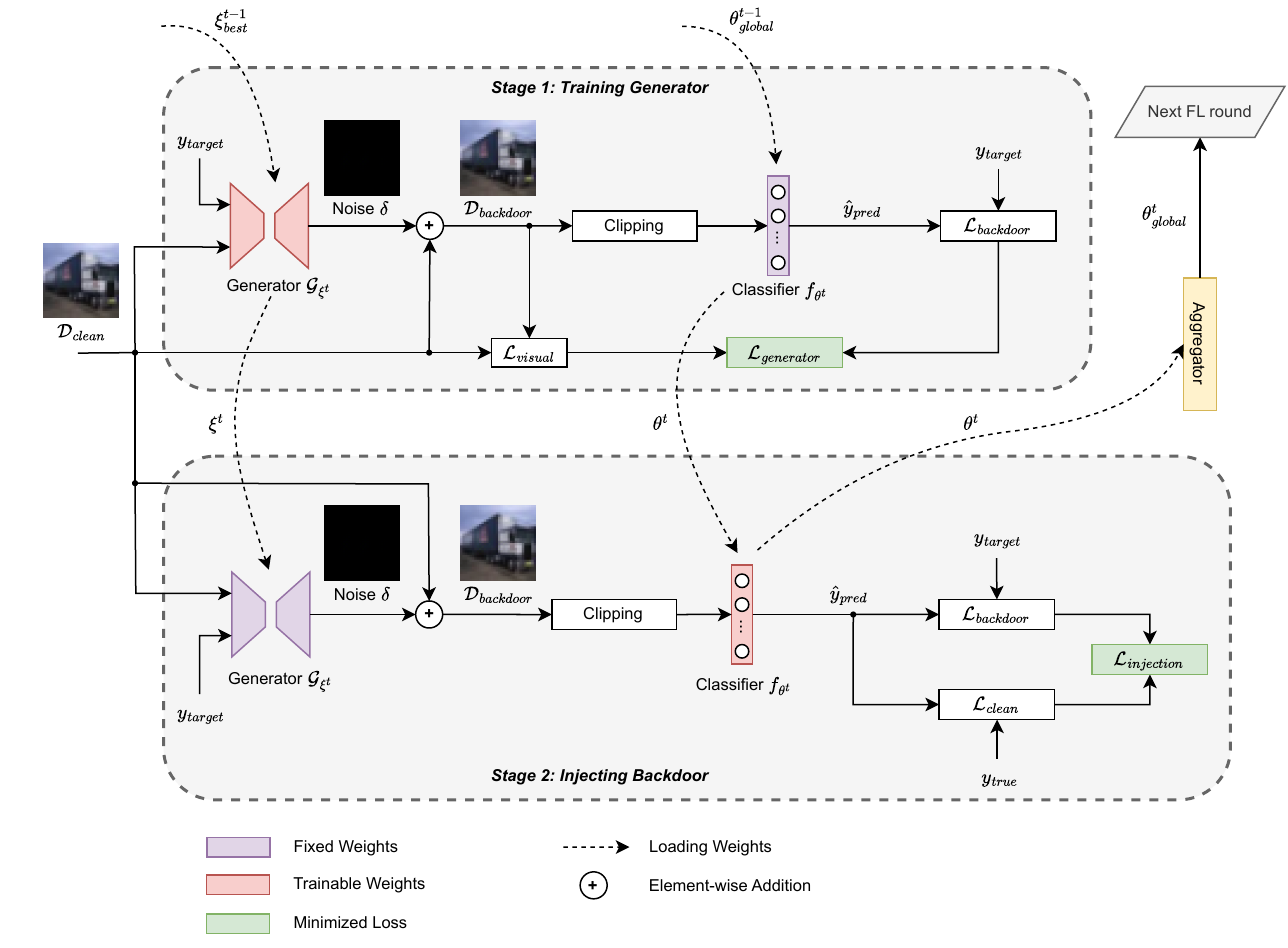}
  \caption{\textbf{Venomancer framework.} Our proposed backdoor attack consists of two training stages: (1) Training generator and (2) Injecting backdoor. In the first stage, the generator $\mathcal{G}_{\xi^t}$ is updated using $\xi_{best}^{t-1}$ to generate the adversarial noise $\delta$.
  The generative model is trained using a combination of $\mathcal{L}_{backdoor}$, which misleads the local model into assigning poisoned images to a selected target class, and $\mathcal{L}_{visual}$ that constrains the similarity between backdoor samples and original images. In the second stage, the local model is trained to perform well on clean samples while incorrectly classifying backdoor data to the target class. After that, the malicious local update is sent back to the central server for aggregation.}
  \label{fig:methodology}
\end{figure}

We consider the supervised learning setting where the attacker or adversary aims to inject the backdoor into the global model. The main goal of the attack is to achieve high backdoor accuracy on the backdoor task while maintaining high clean accuracy on the main task. Specifically, a trigger generator is learned during the local training of malicious clients to produce adversarial noises causing the global model to misclassify the poisoned sample. The attack ensures unaffected performance on the main task and makes the poisoned data visually imperceptible. Moreover, the adversary needs the flexibility to choose any target class during inference time and needs to be robust against defenses in FL. Figure~\ref{fig:methodology} illustrates our Venomancer framework.

\subsection{Attacker's capabilities}
Following threat models in FL from previous works~\cite{bagdasaryan2020backdoor,wang2020attack,xie2019dba,zhang2022neurotoxin,fang2023vulnerability}, we assume that the attacker can compromise one or more clients participating in the FL training to access their training datasets. The attacker can access the received global model's weights and the local updates sent to the central server. Furthermore, of all the output classes of the supervised task that the attacker previously knew, he or she can choose any target classes without knowing all output classes for the task. Moreover, the malicious or compromised clients can participate in the training for every $k$ round, i.e. fixed-frequency attacks~\cite{wang2020attack}.

\section{Proposed method}
\label{sec:method}
\subsection{Target-on-Demand backdoor attack}
In the context of supervised learning, our goal is to train the classifier $f_{\theta}:\mathcal{X} \rightarrow \mathcal{Y}$ that maps an input $x \in \mathcal{X}$ to a corresponding output class $y \in \mathcal{Y}$. The classifier's parameters $\theta$ are optimized through a training dataset $\mathcal{D} = \left \{ \left ( x_{i},y_{i} \right ) \right \}_{i=1}^{\left | \mathcal{D} \right |}$, where $\left | \mathcal{D} \right |$ is the total number of samples in the dataset, with $x_i \in \mathcal{X}$ and $y_i \in \mathcal{Y}$. 

\if
Given the trigger generator $T\left ( x\right )$ used to create a poisoned sample $x_{poison}$ from the clean sample $x$, there are two common types of backdoor settings in practice: all-to-one and all-to-all. In the all-to-one setting, all inputs $x$ patched with the trigger are misclassified as a single predefined target class (a constant label), denoted as $y_{c}$, regardless of their original class $y$:
\begin{equation*} 
  f_{\theta}\left ( T\left ( x \right ) \right )=y_{c}, \hspace{6pt} \forall \left ( x,y \right ) \text{ and } y_{c} \in \mathcal{Y}
\end{equation*}

In the all-to-all setting, the inputs patched with the trigger are misclassified as the target class which relies on the original class of the input $y$:
\begin{equation*} 
  f_{\theta}\left ( T\left ( x \right ) \right )=\left ( y+1 \right )\%\left | \mathcal{Y} \right |, \hspace{6pt} \forall \left ( x,y \right )
\end{equation*}
where $\left |\mathcal{Y} \right |$ is the number of output classes in the supervised task, and $\%$ denotes the modulo operation.  
% where $\mathcal{Y}$ is the set of all possible classes in the supervised task, and $\%$ denotes the modulo operation.

In both settings, the attacker can only use a single predefined target label and not change it during inference time. This makes the attack less powerful and less flexible. At inference, the attacker cannot force the controlled classifier $f$ to misclassify the poisoned sample as any target class that he or she pleases.
\fi

In the target-on-demand setting, the attacker can choose arbitrary target classes during inference time. Particularly, this setting can be formulated as follows:
\[
f_{\theta}\left ( x_{clean}\right ) = y_{true}, \hspace{6pt} f_{\theta}\left ( T\left ( x_{clean}, y_{target} \right ) \right )=y_{target}
\]
% \begin{equation}
% {G^{t + 1}} = {G^t} + \frac{\eta }{n}\mathop \sum \limits_{i = 1}^n ({W_{i}^{t+1} } - {G^t})\tag{2}
% \end{equation}
where $T\left ( x_{clean}, y_{target} \right )$ denotes the trigger function creating backdoor data $x_{backdoor}$ from the clean one $x_{clean} \in \mathcal{D}$ and the selected target class $y_{target} \in \mathcal{Y}$, and $y_{true} \in \mathcal{Y}$ is the ground truth label of the clean data $x_{clean}$.

\subsection{Stage 1: Training generator}
Initially, the trigger generator $\mathcal{G}_{\xi^t}$ assigned the compromised clients' aggregated parameters $\xi^{t-1}_{best}$, while the local model $f_{\theta^t}$ acquires the global model's weights $\theta^{t-1}_{global}$ from the previous round $(t-1)$. At this stage, the generator is trainable, but the local model's weights are fixed. Motivated by Conditional GANs~\citep{mirza2014conditional}, $\mathcal{G}_{\xi^t}$ receives the clean dataset $\mathcal{D}_{clean}$ and a target class $y_{target}$ as inputs to craft adversarial noise $\delta$. An extra embedding layer is employed within the generator to transform $y_{target}$ into a modifiable embedding vector that signifies a trigger pattern tied to a particular target class. This adversarial noise is subsequently merged with clean samples $x_{clean} \in \mathcal{D}_{clean}$, generating the backdoor samples $x_{backdoor} \in \mathcal{D}_{backdoor}$, as shown below:

\[
  \delta=\mathcal{G}_{\xi^t}\left ( x, y_{target}\right ), \hspace{6pt} T_{\xi^t}\left ( x \right )=x+\delta, \hspace{6pt} \forall x \in \mathcal{D}_{clean}
\]
Clipping is needed for $\mathcal{D}_{backdoor}$ to constrain the pixel values within the valid range. The generator is optimized to achieve two objectives: (1) $\mathcal{D}_{backdoor}$ need to be visually indistinguishable from $\mathcal{D}_{clean}$, and (2) $\mathcal{D}_{backdoor}$ are misclassified as $y_{target}$ by $f_{\theta^t}$. 

A conventional technique to ensure the imperceptibility of triggers is to set a limit, $\epsilon > 0$, on the $l_{\infty}$ norm, i.e. $\left | \delta \right |_{\infty} \leq \epsilon$, thus maintaining the trigger's modifications within a small and controlled range. However, Figure~\ref{fig:eps_approach} shows that there is a significant performance degradation in backdoor accuracy to get poisoned images imperceptible to human eyes like our attack. Our findings in FL reveal that this $\epsilon$-bounded method falls short, leading to diminished backdoor effectiveness or poisoned data that are visually detectable. To overcome this drawback, we propose the \textit{visual loss} $\mathcal{L}_{visual}(T_{\xi}(x), x), \forall x\in\mathcal{D}_{clean}$, which enables the generator to adaptively identify critical regions in the original images to perturb the noise into by measuring the pixel-wise differences between clean and backdoor samples. Consequently, the generative model is optimized to create perturbations enhancing the similarity between backdoor samples and their clean counterparts, ensuring the imperceptible poisoned data. Additionally, we use cross-entropy loss $\mathcal{L}_{backdoor}(f_{\theta}(T_{\xi}(x)), y_{target}), \forall x \in \mathcal{D}_{clean}, y_{target} \in \mathcal{Y}$, to assess the backdoor's impact. The overall generator loss, $\mathcal{L}_{generator}$, is then calculated as a combined weighted sum of $\mathcal{L}_{visual}$ and $\mathcal{L}_{backdoor}$:
\begin{equation}
  \mathcal{L}_{generator} = \beta \cdot \mathcal{L}_{backdoor} + \left(1 - \beta\right) \cdot \mathcal{L}_{visual}
\end{equation}
The attacker needs to minimize $\mathcal{L}_{generator}$ to achieve aforementioned two objectives. Finally, the generator is updated using gradient descent:
\begin{equation}
  \xi^t \leftarrow \xi^t-atk\_lr \cdot \nabla\mathcal{L}_{generator}\left ( \theta^t, \xi^t, \mathcal{D}_{backdoor} \right )
\end{equation}

\paragraph{\textit{How to select the best generator?}} Given a number of compromised clients, the attacker selects the best generator $\mathcal{G}_{\xi_{best}}$ that achieves the highest local backdoor accuracy among all the compromised clients at the end of each round. The weights $\xi_{best}$ are then used to initialize the generator of each malicious client in the next round. We observed that this approach stabilizes the training process and makes the backdoor task converge faster.

\subsection{Stage 2: Injecting backdoor} After updating the generator in the first stage, the attacker aims to inject the backdoor into the local model of the malicious client. At this phase, the weights of the generator are fixed and the local model's weights are learnable. The attacker's objective is to make the local model perform normally on the clean samples $\mathcal{D}_{clean}$ while misclassifying the backdoored data $\mathcal{D}_{backdoor}$ as the attacker-chosen target class $y_{target}$. The overall loss $\mathcal{L}_{injection}$ is a combination of the clean loss $\mathcal{L}_{clean}\left ( f_{\theta}\left ( x \right ), y \right ), \forall \left ( x,y \right ) \in \mathcal{D}_{clean}$ (i.e. cross-entropy loss) and the backdoor loss $\mathcal{L}_{backdoor}$:
\begin{equation}
  \mathcal{L}_{injection} = \alpha \cdot \mathcal{L}_{clean} + \left(1 - \alpha\right) \cdot \mathcal{L}_{backdoor}
\end{equation}
This multi-objective task is formulated as follows:
\begin{equation}
  \theta^* = \underset{\theta}{\mathrm{argmin}} \text{ } \mathbb{E}_{\left ( x,y \right ) \sim \mathcal{D}_{clean}} \mathcal{L}_{injection}\left ( f_{\theta}\left ( x \right ),y,f_{\theta}\left ( T_{\xi}\left ( x \right ) \right ),y_{target} \right )
\end{equation}
After finishing the training process, the local model has learned both the clean task and the backdoor task, hence it is poisoned. Next, the client sends its malicious local update back to the central server for aggregation. Finally, the global model is manipulated to misclassify the poisoned data as the attacker-chosen target class $y_{target}$. Following the previous work~\citep{iba}, we choose $\alpha=0.5$ to balance the two objectives. The detailed algorithm for the whole training process of a malicious client is displayed in Appendix~\ref{app:malicious}.

\section{Experiments}
\label{sec:exps}
\subsection{Experimental setup}
\label{exp_setup}
% We conduct our FL experiments using the PyTorch 2.0.0 framework~\citep{pytorch}. All experiments are done on a single machine with 252GB RAM, 64 Intel Xeon Gold 6242 CPUs @ 2.80GHz, and 6 NVIDIA RTX A5000 GPUs with 24GB RAM each. The utilized OS is Ubuntu 20.04.6 LTS.
\subsubsection{Dataset}
We evaluate our method on 4 benchmark datasets: MNIST, F-MNIST, CIFAR-10, and CIFAR-100. MNIST dataset comprises 60,000 training images and 10,000 testing images of handwritten digits, each is a $28\times28$ grayscale image, categorized into 10 classes. The grayscale F-MNIST dataset consists of 60,000 training images and 10,000 testing images of fashion items, also sized at $28\times28$ pixels and distributed across 10 classes. CIFAR-10 contains 50,000 training images and 10,000 testing images across 10 classes, each is a color image of size $32\times32$. CIFAR-100 has the same number of $32\times32$ training and testing images as CIFAR-10 across 100 classes.

\subsubsection{Federated learning setup} 
By default, we set the number of clients $N=100$. At the beginning of each communication round, the server randomly selects $M=10$ clients to participate in the FL training. The global model architecture is ResNet-18~\citep{he2016deep}. 
We simulate a non-i.i.d data distribution with a concentration parameter $\varphi=0.5$ by following previous works~\citep{iba,zhang2022neurotoxin}. Each selected client trains the local model for $2$ epochs using SGD optimizer~\citep{ruder2016overview} with a learning rate of $0.01$. The FL training period lasts for $900$ communication rounds. We train the MNIST, F-MNIST, and CIFAR-10 from scratch. For the CIFAR-100 task, we use a ResNet-18 pre-trained for $60$ epochs. The details of the setup are listed in Appendix~\ref{app:train_details}.
\subsubsection{Method setup} 
The attacker compromises $P$ out of $M$ selected clients and is able to poison their training datasets. $P$ malicious clients are allowed to attack in a fixed frequency of $k$ rounds. By default, we choose $k=1$ and $P=2$. Following previous works~\citep{xie2019dba, zhang2022neurotoxin}, the malicious clients are fixed during the FL training. Each malicious client has a generator that is trained in $5$ epochs using Adam optimizer~\citep{kingma2014adam}. The weight $\beta$ between the visual loss and backdoor loss is $0.01$. In this work, we use an auto-encoder architecture for MNIST, F-MNIST, and CIFAR-10 and an U-Net architecture~\citep{ronneberger2015u} for CIFAR-100. 

We consider the target-on-demand version of the Blend~\citep{chen2017targeted} attack and A3FL~\citep{zhang2024a3fl}, named Blend-ToD and A3FL-ToD respectively, as the baseline attacks. For Blend-ToD, the attack's details are listed in Appendix~\ref{app:baseline_setting}. For A3FL-ToD, we uniformly sample a random target class for each backdoor sample to convert the one-target attack to the multi-target one. The hyperparameters for the selected defenses are detailed in Appendix~\ref{app:defenses}.

\subsubsection{Evaluation metrics}
\begin{itemize}
  \item[\textbullet] \textbf{Clean Accuracy (CA):} We define clean accuracy as the percentage of clean samples (samples without a trigger) that are correctly classified by the global model on the test set.
  \item[\textbullet] \textbf{Backdoor Accuracy (BA):} We define backdoor accuracy as the percentage of poisoned samples that are misclassified as the target class by the global model on the test set. The higher the backdoor accuracy, the more effective the backdoor attack.
  % \item[\textbullet] \textbf{Lifespan}: The lifespan of a backdoor attack is defined as the number of communication rounds during which the attack remains effective. Specifically, the lifespan of a backdoor starts from the round that the attack stops and ends when the BA drops below a certain threshold. Following the previous work \cite{zhang2022neurotoxin}, we choose the threshold as 50\%. A long lifespan indicates that the backdoor attack is more durable, and the attacker can control the global model for a longer time.
\end{itemize}

\subsection{Experimental results}
\label{exp_result}

% This section highlights the effectiveness, robustness, and imperceptibility of our method, Venomancer. We conduct our experiments on selective benchmark datasets in a simulated FL environment. Our results demonstrate that Venomancer can achieve high BAs, high CAs, and remain undetectable by human inspection.

% \begin{table}[b]
%   \centering
%   % \caption{Clean and backdoor accuracy of our method across four FL datasets.}
%   \caption{Venomancer's effectiveness across multiple benchmark datasets}
%   % \resizebox{\textwidth}{!}{%
%     % \begin{tabular}{l|c|c|c|c|c|c}
%     \begin{tabular}{@{}lcccccc@{}}
%       \toprule
%       Dataset & MNIST & F-MNIST & CIFAR-10 & CIFAR-100 \\
%       \midrule
%       CA (\%) & 99.41 & 90.97 & 69.93 & 58.56 \\
%       BA (\%) & 99.52 & 86.28 & 99.66 & 99.46 \\
%       \bottomrule
%     \end{tabular}
%   % }
%   \label{tab:effectiveness}
% \end{table}

\begin{figure}
  \centering
  \includesvg[width=\linewidth]{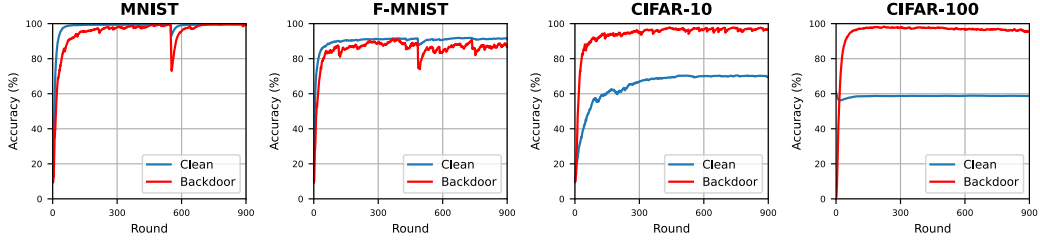}
  % \includegraphics[width=\linewidth]{pdf/high_performance.pdf}
  % \caption{Clean and backdoor accuracy of our method over the training process across four FL datasets.}
  \caption{The effectiveness of Venomancer over the FL training process in selected datasets}
  \label{fig:effectiveness}
\end{figure}

\begin{figure}[ht]
  \centering
  \begin{minipage}[t]{0.45\linewidth}
    \centering
    \includesvg[width=\linewidth]{figures/recent_attacks.svg}
    \caption{Comparison between our method and the target-on-demand version of A3FL and Blend attacks}
    \label{fig:recent_attacks}
    % \vspace{0.5cm}
    % \textbf{Figure 1:} The effectiveness of Venomancer over the FL training process in selected datasets
  \end{minipage}
  \hfill
  \begin{minipage}[t]{0.45\linewidth}
    \centering
    \includegraphics[width=0.95\linewidth]{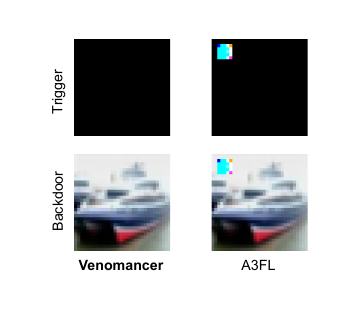}
    \caption{The triggers used by A3FL and our attack}
    \label{fig:triggers_venom_a3fl}
    % \vspace{0.5cm}
    % \textbf{Figure 2:} Clean and backdoor accuracy of our method over the training process across four FL datasets
  \end{minipage}
  % \caption{Comparison of different aspects of the training process in FL datasets}
  % \label{fig:side_by_side}
\end{figure}

\paragraph{Venomancer is more powerful than the target-on-demand version of recent attacks.}Figure~\ref{fig:recent_attacks} demonstrates the performance of our method compared to the A3FL-ToD and Blend-ToD attacks. This experiment runs for 1,000 communication rounds with $\left (P,M,N \right )=\left (4,20,200\right )$. While the CA of considered attacks can be comparable with our attack, i.e. around 60\%, Venomancer outperforms the other two attacks in terms of BA. The Blend-ToD attack has the lowest BA of less 10\%, which is the BA of A3FL-ToD, during the FL period. In contrast, our attack reaches 90\% BA at round $1000^{\text{th}}$.

\paragraph{Venomancer remains effective across the benchmark datasets.}Figure~\ref{fig:effectiveness} shows the performance on the clean task and the backdoor task on the MNIST, F-MNIST, CIFAR-10, and CIFAR-100 datasets after 900 communication rounds from left to right. When the global model converges on the clean task, the BA of our method remains high, exceeding 95\% in MNIST, CIFAR-10, and CIFAR-100 and reaching around 90\% in F-MNIST. The global model accuracies are evaluated on the test set corresponding to each task.

\begin{figure}
  \centering
  \includesvg[width=0.9\linewidth]{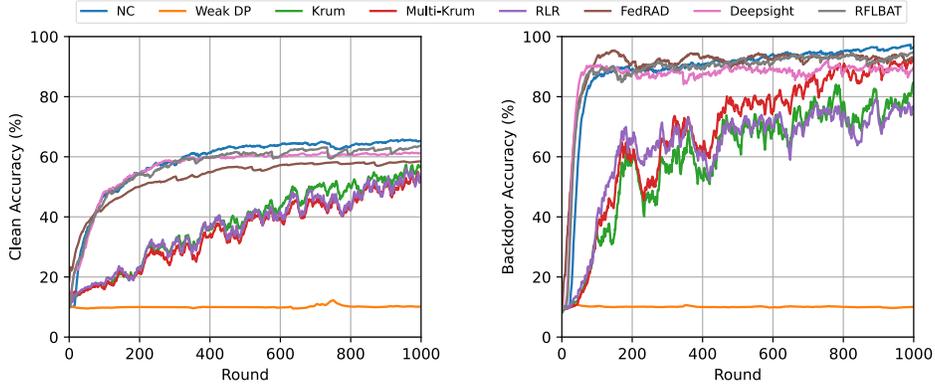}
  % \includegraphics[width=\linewidth]{pdf/high_performance.pdf}
  % \caption{Clean and backdoor accuracy of our method over the training process across four FL datasets.}
  \caption{The robustness of our attack against state-of-the-art-defenses on CIFAR-10}
  \label{fig:robustness}
\end{figure}

% \begin{figure}[t]
%   \centering
%   % \includesvg[width=\linewidth]{figures/eps_approach.svg}
%   \includesvg[width=0.8\linewidth]{figures/mt_vs_venomancer.svg}
%   % \caption{Comparison between the visual loss and the old strategies of using the upper bound $\epsilon$. Our method is both imperceptible and achieves high performance.}
%   \caption{An arbitrary multi-target version of the famous BadNets attack called BadNets-MT is not effective compared to our method on CIFAR-10}
%   \label{fig:mt_vs_venom}
% \end{figure}
% \subsubsection{Performance against existing defenses.}
\paragraph{Venomancer achieves robustness against state-of-the-art defenses.}We empirically demonstrate that Venomancer is robust against the state-of-the-art defenses, as shown in Figure~\ref{fig:robustness} and Table~\ref{tab:robustness}. In this experiment, we choose $\left (P, M, N \right )=\left (4, 20, 200 \right )$. Note that NC stands for Norm Clipping. Table~\ref{tab:robustness} records the inference results after 1,000 communication rounds of training. From Table~\ref{tab:robustness}, we observe that when all defenses (except for Weak DP) are applied, there are slight drops in the clean accuracy compared to the no-defense scenario, i.e. FedAvg, but they keep at reasonable levels, i.e. more than 54\%. For Weak DP, the clean accuracy and backdoor accuracy drop significantly to around 10\%. Most of the other defenses have a minimal impact on the backdoor accuracy, with the backdoor accuracy remaining above 92\% and can be comparable with FedAvg. The RLR defense achieves a backdoor accuracy of 71.58\%. To conclude, the selected defenses are not effective against Venomancer.

% \begin{table}[t]
%   \centering
%   % \caption{Performance against existing defenses. Our method maintains remarkably high backdoor accuracy across four datasets, even when defensive methods are employed.}
%   \caption{Robustness of Venomancer under the selective defenses at the round $900^{th}$}
%   \resizebox{\textwidth}{!}{
%     \begin{tabular}{@{}lcccccccccc@{}}
%       \toprule
%       \multirow{2}{*}{Dataset} & \multicolumn{2}{c}{Norm Clipping} & \multicolumn{2}{c}{Weak DP} & \multicolumn{2}{c}{Krum} & \multicolumn{2}{c}{Multi-Krum} & \multicolumn{2}{c}{No-defense} \\
%       \cmidrule(l){2-3} \cmidrule(l){4-5} \cmidrule(l){6-7} \cmidrule(l){8-9} \cmidrule(l){10-11}
%       & CA (\%) & BA (\%) & CA (\%) & BA (\%) & CA (\%) & BA (\%) & CA (\%) & BA (\%) & CA (\%) & BA (\%) \\
%       \midrule
%       MNIST & 99.45 & 88.83 & 9.80 & 10.08 & 98.06 & 99.88 & 24.41 & 99.92 & 99.41 & 99.52 \\
%       F-MNIST & 91.41 & 88.09 & 89.98 & 90.68 & 87.51 & 98.00 & 88.36 & 99.44 & 90.97 & 86.28 \\
%       CIFAR-10 & 70.62 & 99.13 & 71.44 & 94.91 & 70.16 & 95.19 & 70.42 & 83.20 & 69.93 & 99.66 \\
%       CIFAR-100 & 58.93 & 94.45 & 37.90 & 65.79 & 57.97 & 96.72 & 57.60 & 91.43 & 58.56 & 99.46 \\
%       \bottomrule
%     \end{tabular}
%   }
%   \label{tab:robustness}
% \end{table}

\begin{table}
  \centering
  \caption{Venomancer shows robustness against state-of-the-art defenses on CIFAR-10}
  \resizebox{\linewidth}{!}{%
    \begin{tabular}{@{}lccccccccc@{}}
      \toprule
      Defense & FedAvg & NC & Weak DP & Krum & Multi-Krum & RLR & FedRAD & Deepsight & RFLBAT \\
      \midrule
      CA (\%) & 68.48 & 66.28 & 10.29 & 54.83 & 55.82 & 56.56 & 57.74 & 60.73 & 64.62 \\
      BA (\%) & 88.08 & 98.30 & 10.71 & 86.40 & 89.01 & 71.58 & 93.75 & 92.48 & 97.55\\
      \bottomrule
    \end{tabular}
  }
  \label{tab:robustness}
\end{table}

\begin{figure}
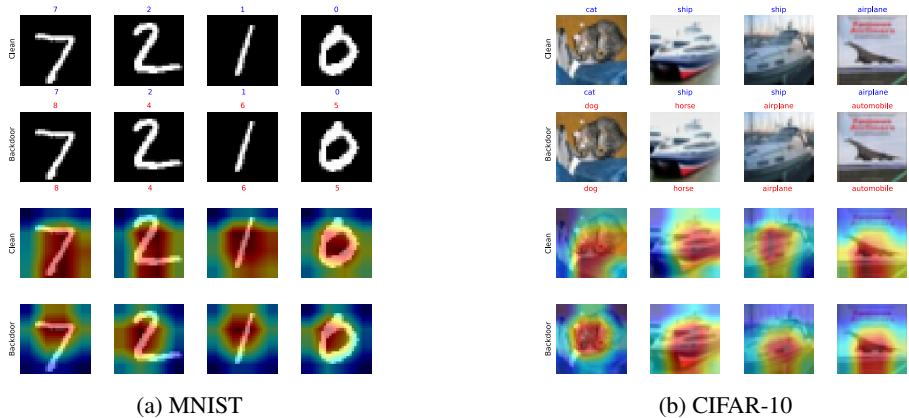

  \centering
  \begin{subfigure}{0.49\linewidth}
    \centering
    \includesvg[scale = 0.25]{figures/imperceptible_mnist_v5.svg}
    \caption{MNIST}
    \label{fig:imperceptible_mnist}
  \end{subfigure}
  \hfill
  \begin{subfigure}{0.49\linewidth}
    \centering
    \includesvg[scale = 0.25]{figures/imperceptible_cifar10_v5.svg}
    \caption{CIFAR-10}
    \label{fig:imperceptible_cifar10}
  \end{subfigure}
  \caption{Imperceptibility of our attack under Grad-CAM heat maps in MNIST and CIFAR-10. The first two rows in (a) and (b) represent the clean and backdoor images in order. The last two rows represent the heat maps of the corresponding clean and backdoor images.}
  \label{fig:high_imperceptibility}
\end{figure}

\begin{figure}
  \centering
  \begin{subfigure}{\linewidth}
    \centering
    \includesvg[width=0.85\linewidth]{figures/eps_bounded.svg}
    \caption{The effectiveness between our attack and the traditional $\epsilon$-bounded approach on CIFAR-10}
    \label{fig:eps_bounded}
  \end{subfigure}
  \vfill
  \begin{subfigure}{\linewidth}
    \centering
    \includegraphics[width=0.9\linewidth]{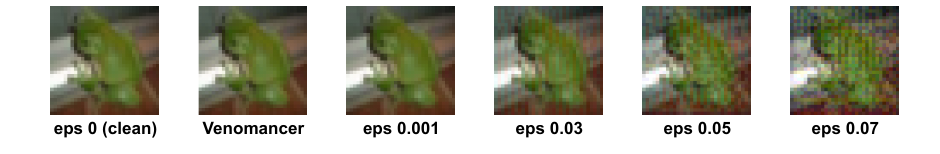}
    \caption{Comparing our poisoned data with other perturbed images corresponding to different $\epsilon$}
    \label{fig:eps_frogs}
  \end{subfigure}
  \caption{Venomancer outperforms the $\epsilon$-bounded approach at the same level of imperceptibility}
  % \caption{Compared with our method, the popular approach of using an upper bound $\epsilon$ on $l_{\infty}$ of adversarial noise shows a trade-off between BA and imperceptibility. }
  \label{fig:eps_approach}
\end{figure}

% \begin{figure}
%   \centering
%   % \includesvg[width=\linewidth]{figures/eps_approach.svg}
%   \includesvg[width=\linewidth]{figures/eps_approach_v8.svg}
%   % \caption{Comparison between the visual loss and the old strategies of using the upper bound $\epsilon$. Our method is both imperceptible and achieves high performance.}
%   \caption{Compared with our method, the popular approach of using an upper bound $\epsilon$ on $l_{\infty}$ of adversarial noise shows a trade-off between BA and imperceptibility. }
%   \label{fig:eps_approach}
% \end{figure}

\paragraph{Venomancer is highly imperceptible to human eyes and target-on-demand.}Figure~\ref{fig:high_imperceptibility} shows the behavior of the backdoored global model on MNIST and CIFAR-10 datasets. The first two rows of each subfigure represent the clean and backdoor images, respectively. The last two rows display the Grad-CAM~\citep{selvaraju2017grad} heat maps of the corresponding clean and backdoor images. The backdoor images are visually imperceptible to human beings and maintain the same visual characteristics as the clean images. The decision-making region in the heat maps for backdoor images is overlapped with that of the clean images. At the top of each image, blue and red represent true classes and target classes, respectively. At the bottom, blue and red show predicted classes for clean and backdoor images, respectively. This result demonstrates the stealthiness and flexibility of our attack.

\paragraph{Venomancer outperforms the traditional $\epsilon$-bounded approach.}To be comparable with our method, a popular approach uses an upper bound $\epsilon$ on the $l_{\infty}$ norm of the adversarial noise to ensure a small perturbation. Figure~\ref{fig:eps_approach} shows that with $\epsilon=0.07$ the BA of the $\epsilon$-bounded approach reaches close to the BA of our attack during the FL training but the corresponding backdoor image is not visually imperceptible. To get the same level of imperceptibility as Venomancer, one needs to decrease $\epsilon$ to $0.001$ but with the trade-off of a BA of only around 40-50\% during the whole FL period.

% To be comparable with our method, a popular approach of using an upper bound $\epsilon$ on the $l_{\infty}$ norm of the adversarial noise to ensure a small perturbation results in unclear artifacts on the original image. Figure~\ref{fig:eps_approach} shows that with $\epsilon=0.07$ the BA of the $l_{\infty}$ constraint approach reaches above but close to the BA of our method during the FL training but the corresponding backdoor image is not visually imperceptible, while the counterpart from our method displays an indistinguishable image from the original one. In order to get the same level of imperceptibility as Venomancer, we can choose $\epsilon=0.001$ but the BA for this case is only around 40-50\% during the whole FL training.

\section{Limitations}
\label{sec:limitation}
Even though our attack shows significant results in terms of imperceptibility and target-on-demand backdoor attacks in FL, it requires training a generative model, e.g. Autoencoder or U-Net, to generate a trigger. This process may be computationally expensive and time-consuming. One possible solution is to use a pre-trained global model because it is easier to inject a backdoor attack into the FL system when the model nearly converges on the clean task but there is no theoretical proof to show that it works. In the future, we plan to address this limitation by exploring more efficient methods to reduce the algorithm complexity without degrading the attack's effectiveness.

% Even though our method shows significant results in terms of imperceptibility and target-on-demand backdoor attacks in FL, it requires the attacker to control a certain number of participated clients to achieve high BA. Moreover, the attacker needs to poison a large number of samples in the training dataset of each malicious client to generate an effective trigger pattern. This may not be feasible in practice, especially when the number of training samples is limited. In the future, we plan to address these limitations by exploring more efficient methods to generate a good trigger with a small number of poisoned samples.

\section{Conclusion}
\label{sec:conclusion}
This work introduces Venomancer, a novel and effective backdoor attack in FL using a generative model with the proposed visual loss to create imperceptible backdoor samples. It bypasses recent defenses while maintaining the performance on benign inputs in the target-on-demand scenario, i.e. allowing attackers to select any target classes during inference. Evaluations on multiple benchmark datasets show the attack's effectiveness in subtly manipulating model predictions. Venomancer highlights the urgent need for better defenses against adaptable backdoor attacks, encouraging the cybersecurity field to develop stronger defenses for trustworthy FL systems.

% In this work, we introduce Venomancer, a novel and effective backdoor attack in FL. This method leverages a generative model with a proposed visual loss to produce imperceptible backdoor samples, effectively bypassing recent defenses while maintaining the model's performance for benign inputs in the target-on-demand scenario, i.e. enabling adversaries to flexibly choose any target classes during inference. Our evaluations across multiple benchmark datasets show our method's effectiveness in manipulating model predictions imperceptibly. The advent of Venomancer highlights the urgent requirement for improved defenses against adaptable backdoor attacks, prompting the cyber-security field to create stronger defense mechanisms for a trustworthy FL system.

\bibliography{neurips_2024}
%%%%%%%%%%%%%%%%%%%%%%%%%%%%%%%%%%%%%%%%%%%%%%%%%%%%%%%%%%%%
\newpage

\appendix

% \section{Appendix}
\section{Algorithms}
\subsection{Local training for a benign client}
\label{app:benign}
The pseudocode for local training of a benign client is shown in Algorithm~\ref{alg:benign}.
\begin{figure}
\vspace*{-\baselineskip}
\begin{minipage}{\columnwidth}
\begin{algorithm}[H]
   \caption{Local training process at round $t$ for a benign client}
   \label{alg:benign}
\begin{algorithmic}
  % \Require Local dataset $\mathcal{D}_{local}$, previous global model's weights $\theta^{t-1}_{global}$, loss function $\mathcal{L}$, local learning rate $lr$, number of local epochs $E$
  % \Ensure Local model's weights $\theta^{t}$
  \Require $\mathcal{D}_{local}$, $\theta^{t-1}_{global}$, $\mathcal{L}$, local learning rate $lr$, number of local epochs $E$
  \Ensure Local model's weights $\theta^{t}$
  % \STATE \textit{Initialize local model $L$ and loss function $l$}:    
  % \INDSTATE[1] $f_{\theta^t} \leftarrow $\theta^t_{global}$ $
  % \INDSTATE[1] $ \ell \leftarrow \mathcal{L}_{class} $
  
  % \State \textbf{function} NormalLocalTraining($\mathcal{D}_{local}$, $\theta^{t-1}_{global}$, $\mathcal{L}$, $lr$, $E$)
  \State \textbf{function} NormalLocalTraining
  \State \hspace{\algorithmicindent} $\theta^t \leftarrow \theta^{t-1}_{global}$ \Comment{Initialize local model} 
  % \State $ \ell \leftarrow \mathcal{L}_{class} $
  \State \hspace{\algorithmicindent} \textbf{for} epoch $e$ $\in$ \{1...$E$\}
      \State \hspace{\algorithmicindent} \hspace{\algorithmicindent} \textbf{for} batch $b$  $\in \mathcal{D}_{local}$
          \State \hspace{\algorithmicindent} \hspace{\algorithmicindent} \hspace{\algorithmicindent}  $\theta^{t} \leftarrow \theta^{t} - lr \cdot \nabla  \mathcal{L} (\theta^{t} , b)$ \Comment{Update local model's weights}
      \State \hspace{\algorithmicindent} \hspace{\algorithmicindent} \textbf{end for}
  \State \hspace{\algorithmicindent} \textbf{end for}
  \State \hspace{\algorithmicindent} \Return $\theta^{t}$
  \State \textbf{end function}
\end{algorithmic} 
\end{algorithm}
\end{minipage}
\end{figure}

\subsection{Local training for a malicious client}
\label{app:malicious}
The pseudocode for local training of a malicious client is shown in Algorithm~\ref{alg:malicious}.
\begin{figure}
  \vspace*{-\baselineskip}
  \begin{minipage}{\columnwidth}
  \begin{algorithm}[H]
     \caption{Local training process at round $t$ for a malicious client}
     \label{alg:malicious}
  \begin{algorithmic}
    % \Require Local dataset $\mathcal{D}_{local}$, previous global model's weights $\theta^{t-1}_{global}$, best previous attacker model's weights $\xi^{t-1}_{best}$, clean loss function $\mathcal{L}_{clean}$, backdoor loss function $\mathcal{L}_{backdoor}$, local learning rate $lr$, attacker's learning rate $atk\_lr$, number of poison epochs $E_{poison}$, injection factor $\alpha$, generator factor $\beta$
    % \Ensure Local model's weights $\theta^{t}$, attacker model's weights $\xi^{t}$
    \Require $\mathcal{D}_{local}$, $\theta^{t-1}_{global}$, $\xi^{t-1}_{best}$, $\mathcal{L}_{clean}$, $\mathcal{L}_{backdoor}$, $y_{target}$, local learning rate $lr$, attacker's learning rate $atk\_lr$, number of poison epochs $E_{poison}$, $\alpha$, $\beta$
    \Ensure Local model's weights $\theta^{t}$, attacker model's weights $\xi^{t}$
    % \STATE \textit{Initialize local model $L$ and loss function $l$}:    
    % \INDSTATE[1] $f_{\theta^t} \leftarrow $\theta^t_{global}$ $
    % \INDSTATE[1] $ \ell \leftarrow \mathcal{L}_{class} $
    
    % \State \textbf{function} BackdoorLocalTraining($\mathcal{D}_{local}$, $\theta^{t-1}_{global}$, $\xi^{t-1}_{best}$, $\mathcal{L}_{clean}$, $\mathcal{L}_{backdoor}$, $\mathcal{L}_{visual}$, \newline\phantom{\textbf{function} BackdoorLocalTraining(}$lr$, $atk\_lr$, $E_{poison}$, $\alpha$, $\beta$)
    \State \textbf{function} BackdoorLocalTraining
    \State \hspace{\algorithmicindent} $\theta^t \leftarrow \theta^{t-1}_{global}$ \Comment{Initialize local model} 
    \State \hspace{\algorithmicindent} $\mathcal{G}_{\xi^t} \leftarrow \xi^{t-1}_{best}$ \Comment{Initialize attacker model} 
    \State \hspace{\algorithmicindent} $\mathcal{D}_{clean} \leftarrow \mathcal{D}_{local}$
    % \State $ \ell \leftarrow \mathcal{L}_{class} $
    \State \hspace{\algorithmicindent} \textbf{for} epoch $e$ $\in$ \{1...$E_{poison}$\}
        \State \hspace{\algorithmicindent} \hspace{\algorithmicindent} \textbf{for} batch $b_{clean}$  $\in \mathcal{D}_{clean}$
            \State \hspace{\algorithmicindent} \hspace{\algorithmicindent} \hspace{\algorithmicindent} $\triangleright$ Stage 1: Training generator
            \State \hspace{\algorithmicindent} \hspace{\algorithmicindent} \hspace{\algorithmicindent} $\delta \leftarrow \mathcal{G}_{\xi^t}\left ( b, y_{target} \right )$ \Comment{Generate adversarial noise}
            \State \hspace{\algorithmicindent} \hspace{\algorithmicindent} \hspace{\algorithmicindent} $b_{poison} \leftarrow \text{Clipping}\left ( b_{clean} + \delta \right )$ \Comment{Create poisoned batch}
            \State \hspace{\algorithmicindent} \hspace{\algorithmicindent} \hspace{\algorithmicindent} $\mathcal{L}_{generator} \leftarrow \beta \cdot \mathcal{L}_{backdoor} + \left(1 - \beta\right) \cdot \mathcal{L}_{visual}$
            \State \hspace{\algorithmicindent} \hspace{\algorithmicindent} \hspace{\algorithmicindent}  $\xi^{t} \leftarrow \xi^{t} - atk\_lr \cdot \nabla  \mathcal{L}_{generator}  (\theta^{t}, \xi^t, b_{poison})$
            \State \hspace{\algorithmicindent} \hspace{\algorithmicindent} \hspace{\algorithmicindent} $\triangleright$ Stage 2: Injecting backdoor
            \State \hspace{\algorithmicindent} \hspace{\algorithmicindent} \hspace{\algorithmicindent} $\mathcal{L}_{injection} \leftarrow \alpha \cdot \mathcal{L}_{clean} + \left(1 - \alpha\right) \cdot \mathcal{L}_{backdoor}$
            \State \hspace{\algorithmicindent} \hspace{\algorithmicindent} \hspace{\algorithmicindent}  $\theta^{t} \leftarrow \theta^{t} - lr \cdot \nabla  \mathcal{L}_{injection}  (\theta^{t}, \xi^t, b_{clean})$
        \State \hspace{\algorithmicindent} \hspace{\algorithmicindent} \textbf{end for}
    \State \hspace{\algorithmicindent} \textbf{end for}
    \State \hspace{\algorithmicindent} \Return $\theta^{t}$, $\xi^{t}$
    \State \textbf{end function}
  \end{algorithmic} 
  \end{algorithm}
  \end{minipage}
\end{figure}

\section{Training details and experimental settings}
\label{app:train_details}
\subsection{Datasets}
\label{app:datasets}
In our experiments, we use 4 benchmark image classification datasets, including MNIST~\citep{deng2012mnist}, Fashion MNIST~\citep{xiao2017fashion} (F-MNIST), CIFAR-10~\citep{krizhevsky2009learning}, and CIFAR-100~\citep{krizhevsky2009learning} to evaluate our method. Those are widely used datasets in the context of backdoor attacks in FL. The usage of a dataset with more classes, i.e. CIFAR-100, enables better evaluation of the scalability of our target-on-demand and imperceptible backdoor attack in FL. The dataset descriptions and used parameters are summarized in Table~\ref{tab:data_description}.

We conduct our FL experiments using the PyTorch 2.0.0 framework~\citep{pytorch}. All experiments are done on a single machine with 252GB RAM, 64 Intel Xeon Gold 6242 CPUs @ 2.80GHz, and 6 NVIDIA RTX A5000 GPUs with 24GB RAM each. The utilized OS is Ubuntu 20.04.6 LTS.

\begin{table}
    \centering
    \caption{Dataset descriptions and FL parameters}
    \resizebox{0.8\textwidth}{!}{
    \begin{tabular}{@{}lcccc@{}}
        \toprule
        & \textbf{MNIST} & \textbf{F-MNIST} & \textbf{CIFAR-10} & \textbf{CIFAR-100} \\
        \midrule
        Classes & 10 & 10 & 10 & 100 \\
        % \midrule
        Image Size & $28\times28\times1$ & $28\times28\times1$ & $32\times32\times3$ & $32\times32\times3$ \\
        % \midrule
        Total Clients & 100 & 100 & 100 & 100 \\
        % \midrule
        Clients/Round & 10 & 10 & 10 & 10 \\
        % \midrule
        Model & ResNet-18 & ResNet-18 & ResNet-18 & ResNet-18 (*) \\
        % \midrule
        Optimizer & SGD & SGD & SGD & SGD \\
        % \midrule
        % $lr/E$ & 0.01$/$2 & 0.01$/$2 & 0.01$/$2 & 0.001$/$2 \\
        % \midrule
        % $atk\_lr/E_{poison}$ & 0.002$/$5 & 0.002$/$5 & 0.002$/$5 & 0.002$/$5 \\
        % \midrule
        % \multirow{2}{*}{Batch size /} & \multirow{2}{*}{64/512} & \multirow{2}{*}{64/512} & \multirow{2}{*}{64/512} & \multirow{2}{*}{64/512} \\
        % Test batch size & & & & \\
        \bottomrule
    \end{tabular}
    }
    \par
    \vspace{2mm} % Adds a little space between the table and the note
    \footnotesize (*) indicates that the model is pre-trained
    \label{tab:data_description}        
\end{table}

\subsection{Detailed experimental settings}
\label{app:detailed_exp}
We follow the FL simulation in~\cite{bagdasaryan2020backdoor, wang2020attack, xie2019dba}, where the central server randomly selects a subset of clients and broadcasts its global model to every local model from participated clients. The selected clients conduct local training on their training dataset for $E$ epochs and then send model updates back to the central server for aggregation, i.e. FedAvg~\cite{mcmahan2017communication}. Inspired by~\citep{bagdasaryan2020backdoor, xie2019dba, iba}, we choose the number of total clients, number of participated clients per FL round, and local epochs $E$ as summarized in Table~\ref{tab:task_specification}.
\begin{table}
    \centering
    \caption{Task specifications and learning parameters}
    \resizebox{\textwidth}{!}{
    \begin{tabular}{@{}lccccc@{}}
        \toprule
        \multirow{2}{*}{Task} & \multirow{2}{*}{Features} & \multirow{2}{*}{$lr/E$} & \multirow{2}{*}{$atk\_lr/E_{poison}$} & \multirow{2}{*}{\makecell{Batch size / \\ Test batch size} } & \multirow{2}{*}{Generator / Attacker's optimizer} \\
        & & & & & \\
        \midrule
        MNIST & 784 & 0.01$/$2 & 0.0002$/$5 & 64$/$512 & Conditional Autoencoder / Adam \\
        % \midrule
        F-MNIST & 784 & 0.01$/$2 & 0.0002$/$5 & 64$/$512 & Conditional Autoencoder / Adam \\
        % \midrule
        CIFAR-10 & 3072 & 0.01$/$2 & 0.0002$/$5 & 64$/$512 & Conditional Autoencoder / Adam \\
        % \midrule
        CIFAR-100 & 3072 & 0.001$/$2 & 0.0002$/$5 & 64$/$512 & Conditional U-Net / Adam \\
        \bottomrule
    \end{tabular}
    }
    \label{tab:task_specification}        
\end{table}

Following~\citep{xie2019dba, wang2020attack}, we simulate the heterogeneous data partition by sampling $p_k \sim Dir_K\left ( \varphi \right )$. Under this strategy, non-i.i.d degree $\varphi=0$ means the data is completely distributed (homogeneity). If $\varphi=1$, the data distribution is absolutely non-i.i.d. To be consistent with prior works~\citep{xie2019dba,wang2020attack}, we select $\varphi=0.5$ in all experiments.

\subsection{Classification model} 
\begin{figure}[t]
  \centering
  \begin{subfigure}{0.49\linewidth}
    % \includesvg[width=\linewidth]{figures/resnet18.svg}
    \includegraphics[width=\linewidth]{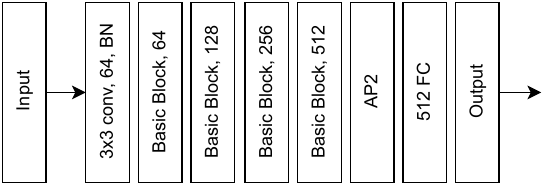}
    \caption{ResNet-18}
  \end{subfigure}
  \hfill
  \begin{subfigure}{0.5\linewidth}
    % \includesvg[width=\linewidth]{figures/resnet19.svg}
    \includegraphics[width=\linewidth]{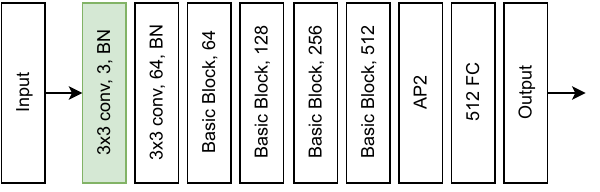}
    \caption{The custom ResNet-18}
  \end{subfigure}
  % \caption{Performance of our method under various defenses. The attack is conducted over the training process.}
  \caption{The ResNet architectures used in our experiments. Green means an additional layer.}
  \label{fig:resnet}
\end{figure}

For MNIST and F-MNIST, we add one convolution layer to the original ResNet-18 architecture to convert grayscale images of $28\times28\times1$ to images of $32\times32\times3$. For CIFAR-10, we use the default ResNet-18 architecture. For CIFAR-100, the ResNet-18 model is pre-trained for 60 epochs because the model learns backdoor features quickly when it is close to vanilla convergence\citep{xie2019dba, bagdasaryan2020backdoor} so the test accuracy for this task starts from $60.94\%$ at the first round of the FL training. For MNIST, F-MNIST, and CIFAR-10, we train the local models from scratch. In ResNet-18, each Basic Block consists of two convolution layers, and the number of channels is doubled every time the spatial size is halved. The detailed architecture for each classifier is shown in Figure~\ref{fig:resnet}.
\subsection{Generator model} 
\begin{figure}
  \centering
  \begin{subfigure}{0.8\linewidth}
    % \includesvg[width=\linewidth]{figures/autoencoder.svg}
    \includegraphics[width=\linewidth]{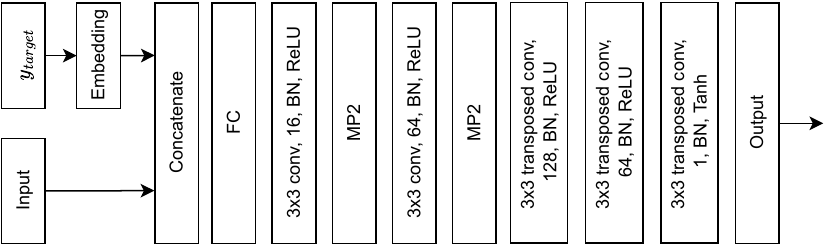}
    \caption{Conditional Autoencoder}
  \end{subfigure}
  \hfill
  \begin{subfigure}{0.9\linewidth}
    % \includesvg[width=\linewidth]{figures/unet.svg}
    \includegraphics[width=\linewidth]{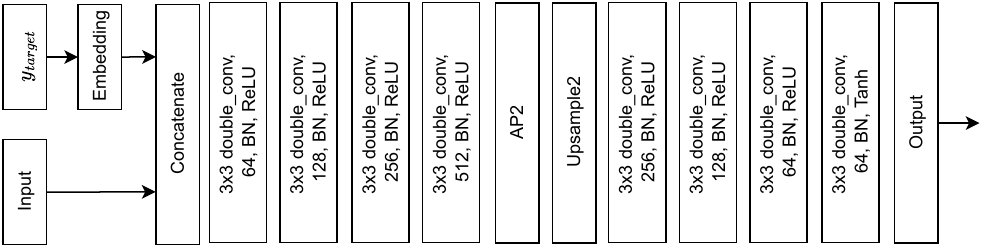}
    \caption{Conditional U-Net}
  \end{subfigure}
  % \caption{Performance of our method under various defenses. The attack is conducted over the training process.}
  \caption{The generator architectures are used in our experiments. The size of a target class embedding vector equals to the number of possible labels, i.e. 10 for MNIST, F-MNIST, and CIFAR-10, and 100 for CIFAR-100.}
  \label{fig:generator}
\end{figure}

In our experiments, we use a simple Conditional Autoencoder and a Conditional U-Net as the generator model. The Conditional Autoencoder is used for MNIST, F-MNIST, and CIFAR-10 while the Conditional U-Net is used for CIFAR-100. For Conditional U-Net, there is a component called \textit{double\_conv} which consists of two convolution layers. The input of the generator is an image and a learnable embedding vector of a target class. The detailed generator architectures are displayed in Figure~\ref{fig:generator}.

\subsection{Detailed setting for Blend-ToD} 
\label{app:baseline_setting}
\begin{figure}[t]
  \centering
  % \includesvg[width=\linewidth]{figures/image_triggers.svg}
  \includegraphics[width=\linewidth]{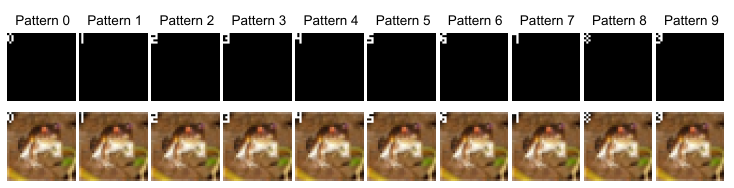}
  \caption{The 10 fixed patterns serve as 10 triggers so that when one pattern is patched onto the top left corner of the image, the global model will predict that image as the trigger-like target class}
  \label{fig:fixed_triggers}
\end{figure}

Inspired by the works from~\citep{chen2017targeted}, we design the baseline attack in FL for the target-on-demand and imperceptible settings to compare with our method. This baseline setting of the multi-target and imperceptible attack is called Blend-ToD. Considering the CIFAR-10 task, we create 10 fixed patterns of shapes 0, 1, 2, 3, 4, 5, 6, 7, 8, 9. Whenever a pattern with the number $i$ is patched onto an image, the global model will predict that image as the target class $i$ ($i\in \{0,1,2,...,9\}$). By default, each fixed trigger is added on the top left corner of the image. We use the following function to generate the image $x$ from the original dataset $\mathcal{D}$ with a fixed trigger $\Delta$ and a blend ratio $\gamma \in \left [ 0,1\right ]$:
\begin{equation}
    x_{poison} = \left ( 1 - \gamma \right ) \cdot x + \gamma \cdot \Delta \cdot mask
\end{equation}
Here, all elements of $mask$ are 0s except for the positions where the trigger is patched. Those positions are 1s. The purpose of the blend ratio $\gamma$ is to control the visibility of the trigger so that the poisoned image is imperceptible to human beings. Inspired by~\citep{chen2017targeted}, we choose $\gamma=0.001$ in training to insert a stealthy trigger to generate poisoned data and do not use the blend ratio during inference to take advantage of the easy-implementability in practice as in the Accessory Injection strategy~\citep{chen2017targeted}. The function used to generate poisoned images without the blend ratio is:
\begin{equation}
    x_{poison} = x + \Delta \cdot mask
\end{equation}
The visualization of the patterns is shown in Figure~\ref{fig:fixed_triggers}.

\subsection{Setting for the $\epsilon$-bounded approach} 
Follow prior works~\citep{iba,Doan_2021_ICCV} on using a small upper bound $\epsilon > 0$ on the adversarial noise in order to make the noise small enough so that the images patched with the noise similar to the original ones, we design experiments to compare with Venomancer by using the formulas below to create poisoned images:
\[
  \delta = \mathcal{G}_{\xi}(x, y_{target}) \cdot \epsilon; \hspace{4pt} x_{poison} = \text{Clip}\left (x + \delta \right )
\]

\section{Hyperparameters for our proposed attack}
\subsection{Controlling coefficient for training generator $\beta$}
When training a generator at stage 1, the generator is trained to achieve two tasks: (1) to generate poisoned images that are imperceptible to human beings and (2) to make the global model predict poisoned images as the arbitrarily attacker-chosen class. The first task is achieved by minimizing the visual loss $\mathcal{L}_{visual}$, i.e. the cosine distance function, between the original image and the poisoned image. The second task is achieved by minimizing the backdoor loss $\mathcal{L}_{backdoor}$, i.e. the cross-entropy loss between the global model's prediction and the target class. Coefficient $\beta$ is used to control the strength of the loss signals from the two tasks. \if In section [cite], we show that the value of $\beta$ has a significant impact on the imperceptibility and the attack effectiveness.\fi In our experiments, we choose $\beta=0.01$ for MNIST, F-MNIST, and CIFAR-10, and $\beta=0.004$ for CIFAR-100.
\subsection{Controlling coefficient for injecting backdoor $\alpha$}
In our studies, the $\alpha$ controls the strength of the clean loss $\mathcal{L}_{clean}$ and the backdoor loss $\mathcal{L}_{backdoor}$, i.e. the cross-entropy loss, during the local training. If $\alpha$ is large, i.e. $\alpha > 0.5$, the classifier's performance on clean data will be quickly converged to the optimum. When $\alpha$ is small, i.e. $\alpha<0.5$, the classifier's performance on backdoor or poisoned data will rapidly achieve a high value. If the generator $\mathcal{G}$ is properly trained, i.e. the learning speed of $\mathcal{G}$ is the same as the learning speed of the classifier, the poisoned classifier still converges to the same optimal performance on both clean and backdoor tasks. So we select $\alpha=0.5$ for all tasks.
\subsection{The learning rate $atk\_lr$ of the generator $\mathcal{G}$}
This parameter controls the learning speed of the generator $\mathcal{G}$ during the local training of a malicious client. Empirically, the learning rate $atk\_lr$ is suggested to be in the range $\left ( 0.0001,0.001\right )$. In our experiments, we choose $atk\_lr=0.0002$ for all tasks.
\subsection{The number of poisoning epochs $E_{poison}$}
In our framework, the trigger generator and the classifier are trained in an alternating manner. First, the generator is trained for $E_{poison}$ epochs, then the classifier is trained for $E$ epochs. Increasing $E_{poison}$ can help the attack to be more effective. However, it also increases the computational cost. In our work, we choose $E_{poison}=5$ for all tasks.
\subsection{The number of malicious clients}
The number of malicious clients is a crucial factor that affects the backdoor accuracy and imperceptibility of the attack. In our experiments, we choose the number of malicious clients to be 2\% of the total clients, e.g. 2 out of 100 total clients, for all tasks.

\section{Hyperparameters for the selected defenses}
\label{app:defenses}
In this paper, we select eight state-of-the-art defenses, i.e. Norm Clipping~\citep{sun2019can} (NC), Weak DP~\citep{sun2019can}, Krum~\citep{blanchard2017machine}, Multi-Krum~\citep{blanchard2017machine}, RLR~\citep{ozdayi2021defending}, FedRAD~\cite{sturluson2021fedrad}, Deepsight~\citep{rieger2022deepsight}, and RFLBAT~\citep{wang2022rflbat} to evaluate the effectiveness of our method. By default, the hyperparameters for the defenses are set up as in the original papers.

\subsection{Norm Clipping} 
Norm Clipping clips the updates from clients when they exceed a certain threshold. This defense effectively constrains the behavior of malicious clients to prevent the global model from being overwhelmed by a small number of clients. The threshold is set to 1 by default.

\subsection{Weak DP} 
Weak DP introduces Gaussian noise $z \sim \mathcal{N}\left ( 0, \sigma^2\right )$ to clients' updates to perturb carefully crafted malicious updates. It should be noted that this mechanism is not tailored for privacy preservation, hence the Gaussian noise is of a smaller magnitude compared to that used in strict differential privacy practices. The standard deviation $\sigma$ is set to 0.002 by default.

\subsection{Krum/Multi-Krum} 
Krum chooses a client or clients whose updates are closest in terms of L2 distance to the updates of other clients. These selected clients are then used to update the global model. While this method is effective at enhancing the model's robustness by excluding many updates, it can also have an impact on the model's accuracy. Multi-Krum is an extension of Krum that selects multiple such clients to update the global model.

\subsection{RLR} 
Robust Learning Rate (RLR) adjusts the learning rate at the aggregation server based on the sign information of updates from clients. By introducing a learning threshold ($\theta$), the server multiplies the learning rate by -1 for dimensions where the sum of the signs of updates is less than $\theta$. This approach aims to maximize the loss in dimensions likely influenced by backdoor attacks, thus reducing the impact of adversarial updates. By default, we choose the learning threshold $\theta=2$.

\subsection{FedRAD} 
FedRAD employs knowledge distillation in a federated learning setting to mitigate backdoor attacks. Instead of aggregating raw model updates, FedRAD distills the knowledge from each client's model to a central global model. This process involves transferring the softened output probabilities (logits) from the client models to the global model, thereby reducing the influence of any single malicious client. We split 10\% of the original training set as the server dataset for knowledge distillation on the server in 2 epochs.

\subsection{Deepsight} 
DeepSight performs deep model inspection to detect and mitigate backdoor attacks in federated learning. It involves analyzing the intermediate representations and activations of models to identify anomalies that could indicate the presence of a backdoor. The approach uses a combination of statistical techniques and machine learning models to differentiate between benign and malicious updates. In our experiments, we set the number of seeds to 3. For each seed value, we generate 20,000 random samples for computing the activations of the neural network's output layer and we fix the threshold $\tau=1/3$.

\subsection{RFLBAT} 
RFLBAT uses Principal Component Analysis (PCA) and K-means clustering to detect and remove backdoored gradients during the aggregation process. The central server reduces the dimension of the gradients using PCA, clusters them with K-means, and identifies the benign clusters based on cosine similarity. It then aggregates the selected benign gradients to form the global model.

\section{Additional experiments for our method}
\subsection{Comparison with the baseline attack Blend-ToD}
\begin{figure}
  \centering
  \includesvg[width=0.95\linewidth]{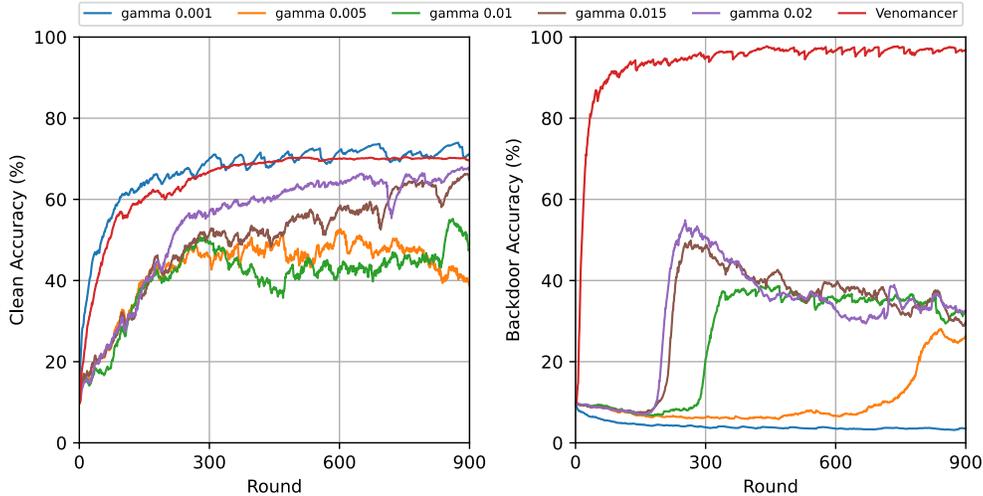}
  \caption{The effectiveness of Venomancer when compared with the baseline attack Blend-ToD for different blend ratios $\gamma$ on CIFAR-10. Our method shows a much higher BA than the baseline at the same level of imperceptibility.}
  \label{fig:baseline}
\end{figure}

% Our attack outperforms BadNets-MT
We compare our method with various blend ratios $\gamma$ of the baseline attack Blend-ToD on CIFAR-10. In this experiment, we extend the FL training to 900 rounds and choose $\gamma$ values for imperceptible triggers. In terms of the Clean Accuracy (CA), Venomancer achieves a CA of around 70\%, which is comparable with the line $\gamma=0.001$. There are fluctuations in the CA with different $\gamma$ values. In Figure~\ref{fig:baseline}, we show that only $\gamma=0.001$, $\gamma=0.015$, and $\gamma=0.02$ reach more than 60\% at the end of the FL training, while the lines for the other $\gamma$ never reaches 60\% during the entire training period. Regarding the Backdoor Accuracy (BA), Venomancer outperforms Blend-ToD for various values of $\gamma$. The BA of Venomancer increases quickly to more than 90\% for the first 200 rounds and remains stable at around 98\% for the rest of the training. In contrast, the BA of Blend-ToD remains at around 10\% during the first 200 rounds. After that, we observe an increase in the BA to around 55\% and 57\% for $\gamma=0.015$ and $\gamma=0.02$ then the BAs decrease to about 40\% and 37\% onwards, respectively. The phenomenon of remaining at a low BA for a while and then increasing to a higher BA is also observed in $\gamma=0.01$ and $\gamma=0.005$. For the one that is comparable with Venomancer's CA, $\gamma=0.001$, its BA remains at below 10\% during the whole training. To conclude, the target-on-demand version of the Blend attack named Blend-ToD, is not effective when the trigger becomes imperceptible. The visualization is shown in Figure~\ref{fig:baseline}.

\subsection{Assessing the imperceptibility of diverse poisoned images}
\begin{figure}
  \centering
  \begin{subfigure}{0.9\linewidth}
    % \includesvg[width=\linewidth]{figures/autoencoder.svg}
    \includegraphics[width=\linewidth]{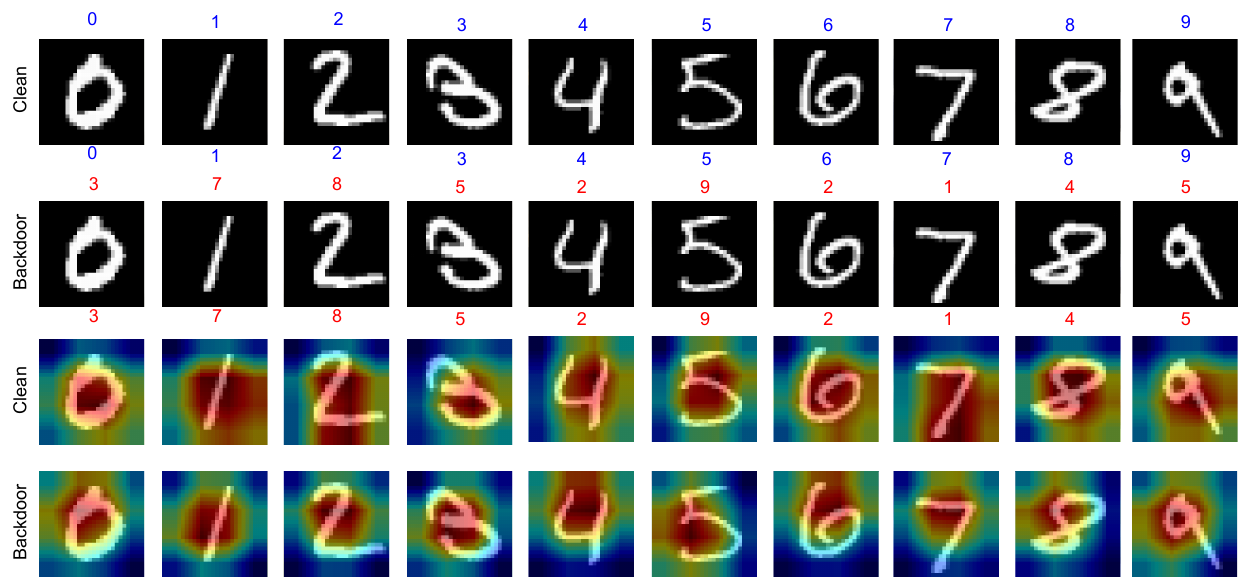}
    \caption{MNIST}
  \end{subfigure}
  \hfill
  \begin{subfigure}{0.9\linewidth}
    % \includesvg[width=\linewidth]{figures/unet.svg}
    \includegraphics[width=\linewidth]{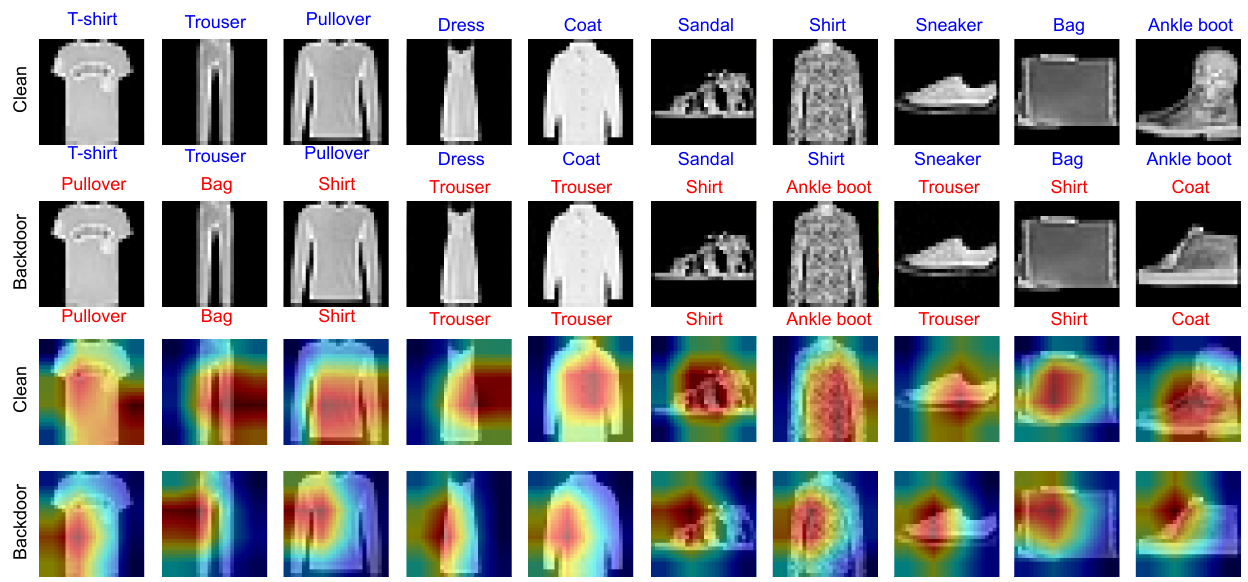}
    \caption{F-MNIST}
  \end{subfigure}
  \hfill
  \begin{subfigure}{0.9\linewidth}
    % \includesvg[width=\linewidth]{figures/unet.svg}
    \includegraphics[width=\linewidth]{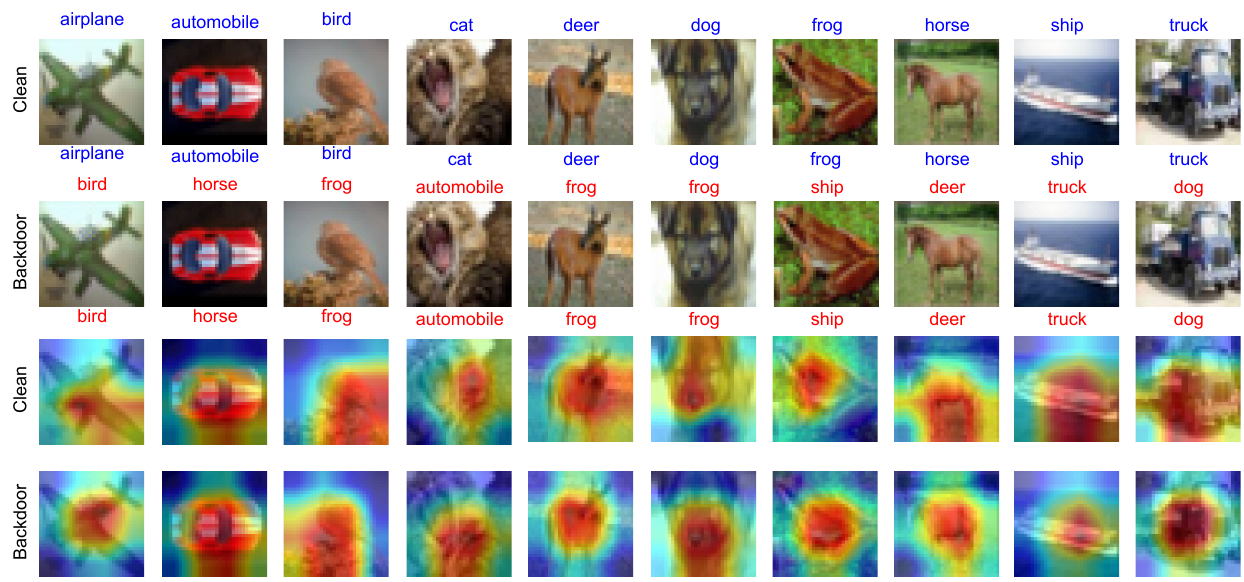}
    \caption{CIFAR-10}
  \end{subfigure}
  % \caption{Performance of our method under various defenses. The attack is conducted over the training process.}
  \caption{The view of clean and backdoor images for MNIST, F-MNIST, and CIFAR-10 on the first two rows, and Grad-CAM~\citep{selvaraju2017grad} visualizations of clean and backdoor images on the last two rows, respectively. Blue and red on top of each image mean the true label and the target class in order. Blue and red under each image mean the predicted label for clean and backdoor images, respectively.}
  \label{fig:imperceptible}
\end{figure}

% Our attack produces nearly perfect imperceptible poisoned images.
% \begin{figure}
%   \centering
%   % \setlength{\abovecaptionskip}{0pt}
%   % \setlength{\belowcaptionskip}{0pt}
%   \includegraphics[width=\linewidth]{figures/cifar10_imperceptible_v2.pdf}
%   \caption{CIFAR-10}
%   \label{fig:cifar10_imperceptible}
% \end{figure}
We visualize the clean and backdoor images for MNIST, F-MNIST, and CIFAR-10 in Figure~\ref{fig:imperceptible}. Our experiments show that the poisoned global model makes a decision for the backdoor image based on the same region as the clean image. Moreover, the backdoor images are impressively visible to human beings, thus emphasizing the stealthiness of our attack. 

\subsection{Backdoor accuracy for each attacker-chosen target class}
Since our method aims to make the poisoned global model misclassify an input patched with a trigger as any attacker-chosen target class during inference without dropping much in the CA, we evaluate the BA of each target class for Venomancer. The results in Table~\ref{tab:ba_each_target_class} show the effectiveness of our attack when it achieves 100\% BA for almost all target classes on MNIST, F-MNIST, and CIFAR-10. Note that for F-MNIST, 0: T-shirt, 1: Trouser, 2: Pullover, 3: Dress, 4: Coat, 5: Sandal, 6: Shirt, 7: Sneaker, 8: Bag, 9: Ankle boot. For CIFAR-10, 0: Airplane, 1: Automobile, 2: Bird, 3: Cat, 4: Deer, 5: Dog, 6: Frog, 7: Horse, 8: Ship, 9: Truck.
\begin{table}
  \centering
  \caption{The BA (\%) of each target class of our method on MNIST, F-MNIST, and CIFAR-10}
  % \resizebox{\textwidth}{!}{%
    \begin{tabular}{@{}lcccccccccc@{}}
      \toprule
      \multirow{2}{*}{Dataset} & \multicolumn{10}{c}{Attacker-chosen target class} \\
      \cmidrule(l){2-11}
      & 0 & 1 & 2 & 3 & 4 & 5 & 6 & 7 & 8 & 9 \\
      \midrule
      MNIST & 100 & 100 & 100 & 100 & 100 & 100 & 100 & 100 & 100 & 100 \\
      F-MNIST & 100 & 99.90 & 100 & 100 & 100 & 89.13 & 100 & 100 & 100 & 98.30 \\
      CIFAR-10 & 100 & 100 & 100 & 100 & 100 & 100 & 100 & 100 & 100 & 100 \\
      \bottomrule
    \end{tabular}
  % }
  \label{tab:ba_each_target_class}
\end{table}

\subsection{Attack effectiveness under different models}
\paragraph{Different classification models.}We evaluate the transferability of the generator to different classification models. In our experiments, we compare VGG11~\citep{simonyan2014very} and ResNet-18. For MNIST, F-MNIST, and CIFAR-10, we train the models from scratch. For CIFAR-100, we use a pre-trained VGG11 and a pre-trained ResNet-18, both of which are trained for 60 epochs.
Our results are shown in Table~\ref{tab:classification_models}.

\begin{table}[h]
  \centering
  % \caption{Performance against existing defenses. Our method maintains remarkably high backdoor accuracy across four datasets, even when defensive methods are employed.}
  \caption{Our method remains effective for both VGG11 and ResNet-18 architectures}
  \resizebox{\textwidth}{!}{
    \begin{tabular}{@{}lcccccccc@{}}
      \toprule
      \multirow{2}{*}{Classifier} & \multicolumn{2}{c}{MNIST} & \multicolumn{2}{c}{F-MNIST} & \multicolumn{2}{c}{CIFAR-10} & \multicolumn{2}{c}{CIFAR-100} \\
      \cmidrule(l){2-3} \cmidrule(l){4-5} \cmidrule(l){6-7} \cmidrule(l){8-9}
      & CA (\%) & BA (\%) & CA (\%) & BA (\%) & CA (\%) & BA (\%) & CA (\%) & BA (\%) \\
      \midrule
      VGG11 & 99.27 & 99.59 & 90.74 & 86.55 & 69.65 & 99.87 & 43.70 & 99.15 \\
      ResNet-18 & 99.51 & 100 & 92.14 & 96.62 & 71.3 & 99.83 & 60.94 & 99.75 \\
      \bottomrule
    \end{tabular}
  }
  \label{tab:classification_models}
\end{table}

\paragraph{Different generator models.}In this section, we evaluate the effectiveness of our method under different generator models, i.e. Conditional Autoencoder and Conditional U-Net. Our results in Table~\ref{tab:generator_models} show that there are not many differences in the BA and CA between the chosen architectures.

\begin{table}[h]
  \centering
  % \caption{Performance against existing defenses. Our method maintains remarkably high backdoor accuracy across four datasets, even when defensive methods are employed.}
  \caption{Our method remains effective for both Conditional Autoencoder and Conditional U-Net}
  \resizebox{\textwidth}{!}{
    \begin{tabular}{@{}lcccccccc@{}}
      \toprule
      \multirow{2}{*}{Generator} & \multicolumn{2}{c}{MNIST} & \multicolumn{2}{c}{F-MNIST} & \multicolumn{2}{c}{CIFAR-10} & \multicolumn{2}{c}{CIFAR-100} \\
      \cmidrule(l){2-3} \cmidrule(l){4-5} \cmidrule(l){6-7} \cmidrule(l){8-9}
      & CA (\%) & BA (\%) & CA (\%) & BA (\%) & CA (\%) & BA (\%) & CA (\%) & BA (\%) \\
      \midrule
      Conditional U-Net & 99.53 & 100 & 92.63 & 99.52 & 74.2 & 100 & 60.94 & 99.75 \\
      Conditional Autoencoder & 99.51 & 100 & 92.14 & 96.62 & 71.3 & 99.83 & 60.70 & 92.20 \\
      \bottomrule
    \end{tabular}
  }
  \label{tab:generator_models}
\end{table}

\subsection{Attack effectiveness under various heterogeneous degrees}
We explore the effectiveness of Venomancer under different $\varphi$ values in the Dirichlet distribution on the four chosen datasets, i.e. MNIST, F-MNIST, CIFAR-10, and CIFAR-100. This highlights the stability and resilience of our method when deployed in conventional FL settings. Table~\ref{tab:heterogenity} presents the results of CA and BA under different $\varphi$ values in the Dirichlet distribution.

\begin{table}[ht]
  \centering
  % \caption{Performance against existing defenses. Our method maintains remarkably high backdoor accuracy across four datasets, even when defensive methods are employed.}
  \caption{The CA and BA of Venomancer under different $\varphi$ values in the Dirichlet distribution}
  \resizebox{\textwidth}{!}{
    \begin{tabular}{@{}lcccccccc@{}}
      \toprule
      \multirow{2}{*}{Task} & \multicolumn{2}{c}{$\varphi=0.2$} & \multicolumn{2}{c}{$\varphi=0.5$} & \multicolumn{2}{c}{$\varphi=0.7$} & \multicolumn{2}{c}{$\varphi=0.9$} \\
      \cmidrule(l){2-3} \cmidrule(l){4-5} \cmidrule(l){6-7} \cmidrule(l){8-9}
      & CA (\%) & BA (\%) & CA (\%) & BA (\%) & CA (\%) & BA (\%) & CA (\%) & BA (\%) \\
      \midrule
      MNIST & 99.57 & 100 & 99.51 & 100 & 99.60 & 100 & 99.52 & 99.99 \\
      F-MNIST & 92.15 & 98.44 & 92.14 & 96.62 & 92.71 & 92.65 & 92.67 & 92.68 \\
      CIFAR-10 & 72.66 & 99.16 & 71.3 & 99.83 & 73.87 & 99.83 & 73.61 & 99.97 \\
      CIFAR-100 & 61.09 & 99.15 & 60.94 & 99.75 & 60.82 & 99.74 & 61.17 & 99.66 \\
      \bottomrule
    \end{tabular}
  }
  \label{tab:heterogenity}
\end{table}

\subsection{Attack effectiveness under different attack frequencies}
We evaluate our attack's effectiveness under different attack frequencies, i.e. $f\in\{1,2,5,10\}$. The results in Table~\ref{tab:attack_frequencies} show that the BA of Venomancer remains high across all datasets, even when the attack frequency is increased. 

\begin{table}[ht]
  \centering
  % \caption{Performance against existing defenses. Our method maintains remarkably high backdoor accuracy across four datasets, even when defensive methods are employed.}
  \caption{The CA and BA of Venomancer under different attack frequencies $f$}
  \resizebox{\textwidth}{!}{
    \begin{tabular}{@{}lcccccccc@{}}
      \toprule
      \multirow{2}{*}{Task} & \multicolumn{2}{c}{$f=1$} & \multicolumn{2}{c}{$f=2$} & \multicolumn{2}{c}{$f=5$} & \multicolumn{2}{c}{$f=10$} \\
      \cmidrule(l){2-3} \cmidrule(l){4-5} \cmidrule(l){6-7} \cmidrule(l){8-9}
      & CA (\%) & BA (\%) & CA (\%) & BA (\%) & CA (\%) & BA (\%) & CA (\%) & BA (\%) \\
      \midrule
      MNIST & 99.51 & 100 & 99.54 & 96.61 & 99.55 & 93.00 & 99.60 & 95.45 \\
      F-MNIST & 92.14 & 96.62 & 92.41 & 91.84 & 92.60 & 90.70 & 92.70 & 90.97 \\
      CIFAR-10 & 71.30 & 99.83 & 74.36 & 100 & 74.01 & 95.81 & 73.76 & 96.50 \\
      CIFAR-100 & 60.94 & 99.75 & 61.97 & 99.57 & 62.09 & 98.85 & 62.09 & 98.46 \\
      \bottomrule
    \end{tabular}
  }
  \label{tab:attack_frequencies}
\end{table}

\subsection{Attack effectiveness under various numbers of malicious clients}
We test our attack with different numbers of malicious clients $P\in\{1,2,5\}$. Our results in Table~\ref{tab:malicious_clients} show that the BA of Venomancer remains effective across all datasets.

\begin{table}[h]
  \centering
  % \caption{Performance against existing defenses. Our method maintains remarkably high backdoor accuracy across four datasets, even when defensive methods are employed.}
  \caption{The CA and BA of Venomancer under different numbers of malicious clients}
  % \resizebox{\textwidth}{!}{
    \begin{tabular}{@{}lcccccccc@{}}
      \toprule
      \multirow{2}{*}{Task} & \multicolumn{2}{c}{$P=1$} & \multicolumn{2}{c}{$P=2$} & \multicolumn{2}{c}{$P=5$} \\
      \cmidrule(l){2-3} \cmidrule(l){4-5} \cmidrule(l){6-7}
      & CA (\%) & BA (\%) & CA (\%) & BA (\%) & CA (\%) & BA (\%) \\
      \midrule
      MNIST & 99.53 & 91.87 & 99.51 & 100 & 99.61 & 100  \\
      F-MNIST & 92.72 & 83.42 & 92.14 & 96.62 & 91.96 & 99.92 \\
      CIFAR-10 & 75.68 & 95.05 & 71.30 & 99.83 & 69.95 & 100 \\
      CIFAR-100 & 61.80 & 99.23 & 60.94 & 99.75 & 57.81 & 99.96 \\
      \bottomrule
    \end{tabular}
  % }
  \label{tab:malicious_clients}
\end{table}

\section{Societal Impact}
\label{app:societal}
Our research aims to enhance recognition and comprehension of vulnerabilities encountered during the training of neural networks in FL settings. If misapplied or in the absence of more robust defenses, the attack we have introduced could pose risks to current FL applications. We view our contributions as crucial for grasping the extent of backdoor attacks within FL settings. We expect that these insights will encourage the advancement of more secure FL frameworks and the creation of stronger defenses.

\end{document}